\documentclass{article}
\usepackage{ijcai26}

\usepackage{times}
\usepackage{soul}
\usepackage{tikz}
\usepackage{url}
\usepackage[hidelinks]{hyperref}
\usepackage[utf8]{inputenc}
\usepackage[small]{caption}
\usepackage{graphicx}
\usepackage{amsmath}
\usepackage{amsthm}
\usepackage{booktabs}
\usepackage{algorithm}
\usepackage{algorithmic}
\usepackage[switch]{lineno}
\usepackage{amsfonts}
\usepackage{blindtext}

\title{Learning long term climate-resilient transport adaptation pathways under direct and indirect flood impacts using reinforcement learning}

\author{
Miguel Costa$^1$\and
Arthur Vandervoort$^1$\and
Carolin Schmidt$^2$\and
Morten W. Petersen$^1$\and
Martin Drews$^1$\and
Karyn Morrissey$^3$\and
Francisco C. Pereira$^1$\\
\affiliations
$^1$Department of Technology, Management and Economics, Technical University of Denmark, 2800 Kgs. Lyngby, Denmark.\\
$^2$School of Management, Technical University of Munich, 80333 Munich, Germany.\\
$^3$J.E. Cairnes School of Business \& Economics, University of Galway, Galway, Ireland.\\
\emails
\{migcos, apiva, \}@dtu.dk,
carolin.schmidt@tum.de, 
\{mwipe, mard\}@dtu.dk,
karyn.morrissey@universityofgalway.ie,
camara@dtu.dk
}

\begin{document}

\maketitle

\begin{abstract}
Climate change is expected to intensify rainfall and other hazards, increasing disruptions in urban transportation systems. Designing effective adaptation strategies is challenging due to the long-term, sequential nature of infrastructure investments, deep uncertainty, and complex cross-sector interactions. We propose a generic decision-support framework that couples an integrated assessment model (IAM) with reinforcement learning (RL) to learn adaptive, multi-decade investment pathways under uncertainty. The framework combines long-term climate projections (e.g., IPCC scenario pathways) with models that map projected extreme-weather drivers (e.g. rain) into hazard likelihoods (e.g. flooding), propagate hazards into urban infrastructure impacts (e.g. transport disruption), and value direct and indirect consequences for service performance and societal costs. Embedded in a reinforcement-learning loop, it learns adaptive climate adaptation policies that trade off investment and maintenance expenditures against avoided impacts. In collaboration with Copenhagen Municipality, we demonstrate the approach on pluvial flooding in the inner city for the horizon of 2024 to 2100. The learned strategies yield coordinated spatial–temporal pathways and improved robustness relative to conventional optimization baselines, namely \emph{inaction} and \emph{random action}, illustrating the framework’s transferability to other hazards and cities. 
\end{abstract}

\section{Introduction}
Climate change is increasing the intensity and frequency of extreme precipitation, raising the risk of pluvial flooding in many cities \cite{ipcc2023climate,dmi2011adaptation}. Urban transport networks are particularly exposed: flooding can reduce speeds, trigger rerouting, and force road and rail closures, with cascading effects on accessibility and economic activity \cite{pregnolato2017climate,wang2020climate}. A salient example is Copenhagen’s 2011 cloudburst, which caused widespread disruption in and around the Danish capital and substantial damages with an estimated 805 million Euros (6 billion Danish kroner, DKK) damage cost \cite{hvass2011sadan,gerdes2012what}. 

Designing climate adaptation strategies is difficult because effective measures are (i) \emph{sequential} and long-lived infrastructure decisions, (ii) evaluated under \emph{deep uncertainty} about future climate conditions, and (iii) embedded in coupled systems where hazard dynamics, network performance, and impacts interact. While prior work has largely focused on short-term operational responses or local interventions, methods for \emph{multi-decade} (i.e., long-term) planning that explicitly optimize adaptive pathways under uncertainty remain limited.

We address this gap by proposing a reinforcement learning (RL) decision-support framework that couples long-term climate projections with hazard modeling, infrastructure and transportation systems simulation, and impact quantification, enabling the learning of adaptive policies over multi-decade horizons. In collaboration with Copenhagen Municipality, we instantiate the framework for an urban case study and evaluate candidate measures (adaptation interventions) across multiple climate scenarios.

Our contributions are threefold:
\begin{itemize}
    \item A generic integrated assessment modeling framework that can be instantiated with modules for climate drivers, hazard processes, infrastructure-system simulation, and direct/indirect impact assessment.
    \item An RL formulation that learns adaptive, long-term sequences of climate adaptation measures under uncertainty; and
    \item An empirical evaluation across climate scenarios to compare robustness and resulting adaptation pathways.
\end{itemize}

% ==================================================== %
% Background
% ==================================================== %
\section{Background}
\label{sec:background}

\subsection{AI for climate adaptation}
Climate adaptation planning requires selecting sequences of interventions under deep uncertainty\footnote{We refer to \emph{deep uncertainty} as situations where analysts do not know, or stakeholders cannot agree on, the appropriate models linking actions to consequences, the probability distributions for key uncertainties, and/or how to value and trade off outcomes \cite{marchau2019dmdu}.}, where impacts emerge over long horizons and propagate through coupled urban systems (e.g., water, transport, energy, communications). Evaluating candidate strategies therefore demands models that (i) translate climate drivers into hazard likelihoods and intensities, (ii) propagate hazards into systems' performance and infrastructure and service disruption, and (iii) value direct and indirect consequences in a way that supports comparison of trade-offs.

Artificial Intelligence (AI) methods are increasingly proposed as decision-support tools for this setting because they can integrate heterogeneous models, capture nonlinear interactions, and optimize policies under uncertainty and long horizons \cite{rolnick2022tackling,cheong2022artificial,tyler2023ai}. Reinforcement learning is particularly relevant when decisions are sequential and outcomes are delayed, enabling policies that balance short- and long-term objectives \cite{matsuo2022deep,gilbert2022choicesrisksrewardreports}. 

In our instantiation, we focus on pluvial flooding and urban transport: flooding disrupts transport through direct damage to infrastructure and operational effects such as reduced speeds, rerouting, and closures, with downstream accessibility and economic impacts \cite{markolf2019transportation,lu2022overview,wang2020climate}. Quantifying both direct and indirect transport impacts is therefore central for evaluating adaptation options and conducting cost-benefit analyses \cite{chang2011future,shahdani2022assessing,ding_interregional_2023}. In flood-related settings, RL has been used mainly for reactive control and response (e.g., drainage/stormwater control, emergency routing) \cite{tian2023flooding,bowes2021flood,li2024reinforcement,fan2021evaluating}. Work on \emph{proactive, multi-decade} flood-risk planning with RL remains limited; for instance, \cite{fend2025reinforcement} studies coastal flood risk management over 2000--2100 but with coarse decision intervals (e.g., decisions every decade), which constrains adaptation flexibility. We target the complementary problem of learning long-horizon, fine-grained \emph{adaptation pathways} that couple hazard evolution with infrastructure-system impacts.

\subsection{Integrated assessment models}
Integrated assessment models (IAMs) couple representations of the climate system with human and technological systems (e.g., energy, land use, macroeconomy) to explore long-run consequences of alternative assumptions and policy choices. They have been widely used to produce internally consistent scenario ensembles and to inform Intergovernmental Panel on Climate Change (IPCC) assessments, but their scale and structural complexity often make them computationally intensive and difficult to embed in iterative decision-optimization workflows \cite{xiong2025emiam}.  

A growing literature therefore uses AI and machine learning (ML) to increase the accessibility and decision-relevance of IAMs. One line of work builds \emph{emulators} (surrogates) that approximate IAM behavior at a fraction of the computational cost, enabling rapid exploration of scenario spaces and sensitivity analyses. Recent examples include emulator frameworks based on marginal abatement cost curve representations of multiple IAMs \cite{xiong2025emiam} and ML-based emulation trained on large scenario databases to reproduce IAM outputs across model families \cite{shin2026mliam}. In parallel, macro-climate modeling reviews emphasize that advances in modern numerical methods and deep learning are increasingly used to solve and analyze richer dynamic stochastic IAMs that would otherwise be intractable \cite{fernandezvillaverde2024climate}.

Beyond acceleration, AI methods support (i) calibration and uncertainty analysis of high-dimensional model components, and (ii) closed-loop decision support, where policies are optimized against long-horizon, uncertain dynamics. Taken together, these developments motivate IAM-based environments that remain modular—so domain models can be swapped or refined—while being computationally amenable to learning and optimization methods.

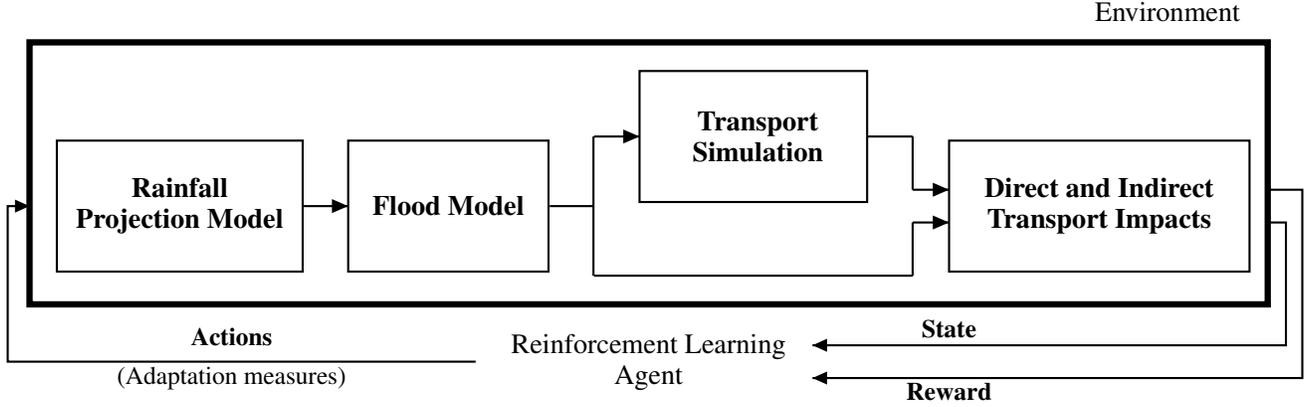
\begin{figure*}[htb]
    \centering
    \resizebox{0.98\linewidth}{!}{%
        \tikzset{every picture/.style={line width=0.75pt}} %set default line width to 0.75pt        
        \begin{tikzpicture}[x=0.65pt,y=0.65pt,yscale=-.9,xscale=1]

%Shape: Rectangle
\draw [line width=2.25] (25,60) -- (705,60) -- (705,220) -- (25,220) -- cycle  ;

% Text Node
\draw (365,255) node   [align=left] {\begin{minipage}[lt]{120pt}\setlength\topsep{0pt}
\begin{center}
Reinforcement Learning Agent
\end{center}
\end{minipage}};

%Straight Lines [id:da5897083012926433] 
\draw (705,170) -- (715,170) -- (715,245) -- (455,245) ;
\draw [shift={(455,245)}, rotate = 360] [fill={rgb, 255:red, 0; green, 0; blue, 0 }  ][line width=0.08]  [draw opacity=0] (8,-4) -- (0,0) -- (8,4) -- cycle ;
%Straight Lines [id:da6778049630294789] 
\draw (705,150) -- (724,150) -- (724,265) -- (455,265) ;
\draw [shift={(455,265)}, rotate = 360] [fill={rgb, 255:red, 0; green, 0; blue, 0 }  ][line width=0.08]  [draw opacity=0] (8,-4) -- (0,0) -- (8,4) -- cycle ;
%Straight Lines [id:da507302552582728] 
\draw (270,255) -- (13,255) -- (13,160) -- (20,160) ;
\draw [shift={(25,160)}, rotate = 180] [fill={rgb, 255:red, 0; green, 0; blue, 0 }  ][line width=0.08]  [draw opacity=0] (8,-4) -- (0,0) -- (8,4) -- cycle ;

% Text Node
\draw (530,235) node [font=\small] [align=left] {\textbf{State}};
% Text Node
\draw (530,273) node [font=\small] [align=left] {\textbf{Reward}};
% Text Node
\draw (136,240) node [font=\small] [align=left] {\begin{minipage}[lt]{32.83pt}\setlength\topsep{0pt}
\begin{center}
\textbf{Actions}
\end{center}
\end{minipage}};
% Text Node
\draw (136,265) node [font=\small] [align=left] {\begin{minipage}[lt]{100.pt}\setlength\topsep{0pt}
\begin{center}
{\footnotesize (Adaptation measures)}
\end{center}

\end{minipage}};
% Text Node
\draw (650,41) node [align=left] {Environment};

% Text Node
\draw    (40,120) -- (175,120) -- (175,200) -- (40,200) -- cycle  ;
\draw (107.5,160) node   [align=left] {\begin{minipage}[lt]{80pt}\setlength\topsep{0pt}
\begin{center}
\textbf{Rainfall Projection Model}\\{}
\end{center}
\end{minipage}};

% Text Node
\draw    (200,120) -- (310,120) -- (310,200) -- (200,200) -- cycle  ;
\draw (255,160) node   [align=left] {\begin{minipage}[lt]{60pt}\setlength\topsep{0pt}
\begin{center}
\textbf{Flood Model}\\
{}
\end{center}
\end{minipage}};

% Text Node
\draw    (360,77.5) -- (485,77.5) -- (485,157.5) -- (360,157.5) -- cycle  ;
\draw (425,117.5) node   [align=left] {\begin{minipage}[lt]{80pt}\setlength\topsep{0pt}
\begin{center}
\textbf{Transport Simulation}\\
{}
\end{center}
\end{minipage}};

% Text Node
\draw    (530,120) -- (695,120) -- (695,200) -- (530,200) -- cycle  ;
\draw (612.5,160) node   [align=left] {\begin{minipage}[lt]{103pt}\setlength\topsep{0pt}
\begin{center}
\textbf{Direct and Indirect Transport Impacts}\\
{}
\end{center}
\end{minipage}};

% Connection
\draw    (175,160) -- (200,160) ;
\draw [shift={(200,160)}, rotate = 180] [fill={rgb, 255:red, 0; green, 0; blue, 0 }  ][line width=0.08]  [draw opacity=0] (8.93,-4.29) -- (0,0) -- (8.93,4.29) -- cycle ;
% Connection
\draw    (335,117.5) -- (360,117.5) ;
\draw [shift={(360,117.5)}, rotate = 180] [fill={rgb, 255:red, 0; green, 0; blue, 0 }  ][line width=0.08]  [draw opacity=0] (8.93,-4.29) -- (0,0) -- (8.93,4.29) -- cycle ;

% Connection
\draw  (485,117.5) -- (510,117.5) ;
% Connection
\draw  (510,117.5) -- (510,150) ;
% Connection
\draw  (510,150) -- (530,150) ;
\draw [shift={(530,150)}, rotate = 180] [fill={rgb, 255:red, 0; green, 0; blue, 0 }  ][line width=0.08]  [draw opacity=0] (8.93,-4.29) -- (0,0) -- (8.93,4.29) -- cycle ;

% Connection
\draw   (335,202.5) -- (510,202.5) ;
% Connection
\draw   (510,170) -- (510,202.5) ;
% Connection
\draw  (510,170) -- (530,170) ;
\draw [shift={(530,170)}, rotate = 180] [fill={rgb, 255:red, 0; green, 0; blue, 0 }  ][line width=0.08]  [draw opacity=0] (8.93,-4.29) -- (0,0) -- (8.93,4.29) -- cycle ;
§
% Connection
\draw   (327,117.5) -- (327,202.5) ;

% Connection
\draw    (303,160) -- (327,160) ;
            
        \end{tikzpicture}
    }
    \caption{Integrated Assessment Model using reinforcement learning to find the best sequence of transport-related adaptation policies for rainfall events in Copenhagen from 2024--2100.}
    \label{fig:met_iam}
\end{figure*}

% ==================================================== %
% Methods
% ==================================================== %

\section{Problem formulation and IAM environment}
\label{sec:methods}

We study long-horizon climate adaptation planning as a sequential decision problem in which a decision-maker selects spatially distributed interventions over multiple decades and evaluates them under uncertain climate-driven hazards. We implement the planning problem as an Integrated Assessment Model environment (Fig.~\ref{fig:met_iam}) that couples (i) climate-driven forcing and hazard generation (e.g., sampling extreme-weather drivers from scenario-based projections), (ii) event-based physical impact modeling, (iii) infrastructure-system performance simulation (here, multi-modal transport), and (iv) valuation of direct and indirect consequences and costs of interventions. All code and experiments are publicly available\footnote{The anonymous link to our code is provided at https://anonymous.4open.science/r/8EC9. If accepted, we will our code and its documentation publicly available on GitHub.}. At each decision step (e.g., a simulated year), the environment generates a hazard realization from the scenario-conditioned distribution, propagates it through the impact and system modules to obtain disruption outcomes, and returns a state representation and scalar objective signal; this closed loop enables stress-testing policies across alternative climate scenarios within a common simulation backbone.

The resulting sequential decision process is modeled as an Markov Decision Process (MDP) $\langle \mathcal{S}, \mathcal{A}, \mathcal{P}, r, \gamma\rangle$. At step $t \in \{1,\dots,T\}$, the agent observes $s_t \in \mathcal{S}$, selects $a_t \in \mathcal{A}$, transitions to $s_{t+1}\sim \mathcal{P}(\cdot \mid s_t,a_t)$ through the IAM dynamics, and receives a reward $r_t=r(s_t,a_t)$. Rainfall (more generally, the extreme-weather driver) is treated as exogenous; conditional on the scenario and time slice, events are sampled independently. Altogether, the objective is to maximize the long-term cumulative reward, or, more generally
\begin{equation}
\max_{\pi}\ \mathbb{E}_{\pi}\Big[\sum_{t=1}^{T}\gamma^{t-1}r_t\Big],
\end{equation}
with discount factor $\gamma \in (0,1]$.

\paragraph{State space $\mathcal{S}$.}
To capture spatial dependence in flood and transport disruption, we model our representation of the IAM as a graph $G=(V,E)$ with nodes $i\in V$ (traffic-assignment zones) and edges $(i,j)\in E$ (spatial adjacency).
The state at time $t$ is $\{\mathbf{x}_{i,t}\}$ with
\[
\mathbf{x}_{i,t} = \big[I_{i,t},\, D_{i,t},\, C_{i,t},\, \mathbf{z}_{i,t}\big],
\]
where $I_{i,t}$, $D_{i,t}$, and $C_{i,t}$ are IAM-derived impacts (described below) at step $t$ (typically one year): direct infrastructure damage, travel delays, and trip cancellations.
The vector $\mathbf{z}_{i,t}$ encodes the remaining effectiveness of implemented interventions, which scales their volumetric benefit (reduction in water volume, m$^3$) and decays over time since deployment (i.e., to model temporal effectiveness reduction).

\paragraph{Action space $\mathcal{A}$.}
Actions are implemented at the zone level. At each timestep $t$, the agent selects, for each zone $i\in V$, one intervention from a discrete set\\
$a_{i,t}\in\left\{
\begin{aligned}
&\texttt{DoNothing},\ \texttt{Bioretention Planters},\\
&\texttt{Soakaway},\ \texttt{Storage Tank},\ \\ 
&\texttt{Porous Asphalt},\ \texttt{Pervious Concrete}, \ \\
&\texttt{Permeable Pavers},\ \texttt{Grid Pavers} 
\end{aligned}
\right\}.$
yielding the joint action $a_t=\{a_{i,t}\}_{i\in V}$. Once an intervention is deployed in a zone, it remains active for its defined lifetime, and its effectiveness decays over time. To enforce feasibility, interventions that are currently active in zone $i$ are rendered unavailable via action masking, i.e., the policy is restricted to the subset of not implemented actions in that zone. Each possible action has different properties (drainage capacity, lifetime duration, conditions where it can be applied, implementation and maintenance costs) and effects, resulting in different sets of trade-offs between what, where, and when actions can be applied.

\paragraph{Reward.}
We optimize a monetized objective that trades off avoided flood impacts against investment and maintenance. At each step $t$,
\begin{equation}
r_t = -\sum_{i\in V}\Big(I_{i,t} + D_{i,t} + C_{i,t} + A_{i,t} + M_{i,t}\Big),
\label{eq:reward}
\end{equation}
where $A_{i,t}$ is the investment cost incurred by newly deployed interventions at $t$ in zone $i$, and $M_{i,t}$ is the maintenance cost of interventions currently active in zone $i$.

\paragraph{Graph policy parameterization.}
We instantiate $\pi_\theta$ as a message-passing graph neural network (GNN) that maps node features to per-zone action distributions. 
Given the state $s_t=\{\mathbf{x}_{i,t}\}_{i\in V}$, the network performs $L$ rounds of neighborhood aggregation and produces per-node logits $\ell_{i,t}\in\mathbb{R}^{|\mathcal{A}_i|}$:
\[
\{\mathbf{x}_{i,t}\}_{i\in V}\;\xrightarrow{\;\text{GNN}\;}\;\{\ell_{i,t}\}_{i\in V}, 
\]
\[ 
\pi_\theta(a_{i,t}\mid s_t)=\operatorname{Softmax}\!\big(\ell_{i,t}+m_{i,t}\big),
\]
where $m_{i,t}$ is an action-feasibility mask (set to $-\infty$ for infeasible actions). 
This architecture is permutation-invariant to zone ordering, shares parameters across nodes, and transfers to graphs with different sizes and topologies.

\paragraph{Rainfall projection and flood model.}
In our IAM, we use a rainfall projection model to simulate rain events for the 2024--2100 period. Using projected daily rainfall statistics for three climate scenarios (RCP2.6, RCP4.5, RCP8.5) \cite{vanvuuren2011} from the Danish Meteorological Institute's Climate Atlas \cite{dmi2023klimaatlas}, we build the associated rainfall cumulative distribution function. Then, we sample one rainfall event per timestep, represented as the total accumulated daily rainfall, for which we model the associated flooding in Copenhagen. Depending on rainfall intensity, floods can range from minor, recurring events (e.g., nuisance floods) to high-impact, major flooding events (e.g., those resulting from cloudbursts). To model all flood types, we use SCALGO Live~\cite{scalgo}, a simplified, event-based tool for watershed delineation and flood depth modeling. Here, a uniform rainfall distribution of unspecified duration over the study area is assumed, meaning water accumulates simultaneously at all locations. Water is distributed according to terrain characteristics, filling topographic depressions and allowing us to assign water heights to specific transport network locations.

% ======================================== %
% Transport simulation
\paragraph{Transport simulation}

We extract the drivable, cyclable, and walkable transport networks (i.e., roads, cycling lanes, sidewalks) covering Copenhagen's inner city (\textit{indre by}) using \textsc{osmnx} \cite{boeing2024modeling} from OpenStreetMap\footnote{Available at: \url{https://www.openstreetmap.org/}} (OSM, as of July 7, 2024). Next, to capture the impacts of flooding on transport, we model trips made by Copenhagen residents. For this, we start by using data from the Danish National Travel Survey (Transportvaneundersøgelsen, TU) \cite{christiansen2021tu}. Our study area is divided into 29 traffic assignment zones (TAZs, each covering a few city blocks), corresponding to those used in the Danish National Transport Model (Grøn Mobilitetsmodel, GMM) \cite{vejdirektoratet2022gmm}. To model individual trips (~84k total trips), we randomly sampled an origin and destination (OD) location within the respective OD TAZs.

Finally, we simulate the route taken by each individual using their assigned transport mode from origin to destination on the corresponding transport network. We compute the shortest path in terms of travel time for each trip's origin-destination pair. To account for the effect of water levels on travel, we apply depth-disruption functions \cite{pregnolato2017impact,finnis2008field} that map water depths on each segment (road, street, cycling lane, sidewalk) to reduced travel speeds and, therefore, increased travel times, or, in the event of trips having no traversable path from its origin to destination, to cancellation of trips.

% ======================================== %
% Infrastructure and Transportation Impacts
\paragraph{Transportation impacts}
Finally, we compute direct and indirect transportation impacts. Three types of impacts are estimated: 1) direct network infrastructure damage, 2) indirect travel delays, and 3) indirect trip cancellations. 
% ================================
% Infrastructure damage impacts
When a flood occurs, infrastructure is damaged and ought to be repaired. Following an initial estimate of the construction costs for each network segment (following the approach by \cite{ginkel2021flood}), we compute the estimated damage using depth-damage functions \cite{ginkel2021flood}. These map the percentage of damage on a network link depending on its water depth to the economic cost (i.e., repair, cleaning, and resurfacing works) of repairing such damage. Finally, we aggregate these costs as \(I_{i}\) for the \(i\)-th TAZ. 
% ================================
% Travel delay impacts
Next, we compute travel delays for all simulated trips. As water levels rise, travel speed reduces, resulting in travel time delays. At the same time, as some roads begin to get flooded, routes might change because individuals would choose an alternative and faster route. After comparing the original (no flood) travel time with the flood-affected travel time, we compute the associated travel delay and convert it into economic losses using the Danish value of time for travel delays \cite{transportministeriet2022enhedspriser}. Travel delays costs are aggregated as \(D_{i}\) for each TAZ.
% ================================
% Trip cancellation impacts
finally, when there is no traversable path between a trip's origin and destination due to water levels being too high, we consider trips to not occur and result in an abandoned/cancelled trip. Following existing approaches \cite{hallenbeckTravelCostsAssociated}, we estimate these cancelled trips economic losses and aggregate them at the TAZ-level as \(C_i\)

% ==================================================== %
% Results
% ==================================================== %
\section{Case Study: Copenhagen's Inner City}
\subsection{Experimental setup}
We implement the IAM as a Gymnasium environment \cite{towers_gymnasium_2023} in Python and train policies with Stable-Baselines3 \cite{stable_baselines3}. The case study is Copenhagen’s inner city, represented by 29 TAZ nodes where one adaptation action per zone can be selected at each decision step, over a planning horizon of 2024--2100. We train the graph policy using Proximal Policy Optimization (PPO) \cite{schulman2017proximal,huang2020closer} with batch size 64, 1024 environment steps per update, 10 optimization epochs per update, entropy coefficient 0.01, and a KL-divergence limit of 0.2 to constrain policy updates. Training uses 10 parallel environments and runs for up to 4.5 million environment steps, with early stopping when the return plateaus. Unless stated otherwise, we report results over 10 random seeds (mean $\pm$ standard deviation).

% ======================================== %
\begin{figure*}[!ht]
    \centering
    \includegraphics[width=.8\textwidth]{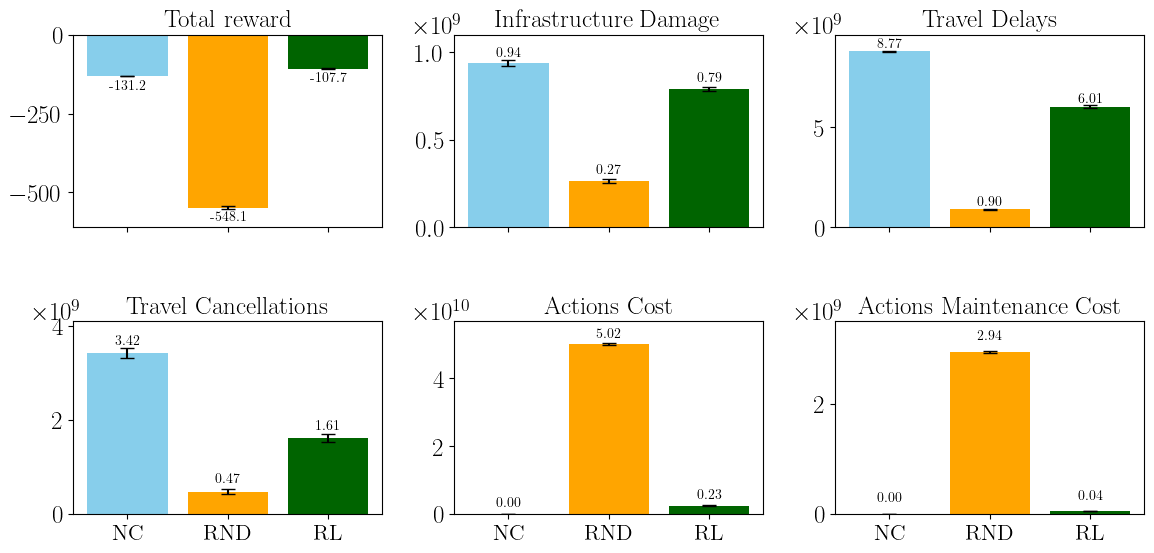} \\
    \vspace{-10pt}
    \caption{Comparison of different adaptation strategies. Total reward (top-left, scaled by \(10^8\)), and reward components are shown: infrastructure damage costs (top-middle), travel delays (top-right), travel cancellations (bottom-left), actions direct costs (bottom-middle), and action maintenance costs (bottom-right). All values correspond to Danish Kroner (DKK). Please note the different y-axis scale across figures.}
    \label{fig:benchmark_algorithms}
\end{figure*}

\begin{figure*}[!ht]
    \centering
    \includegraphics[width=.8\textwidth]{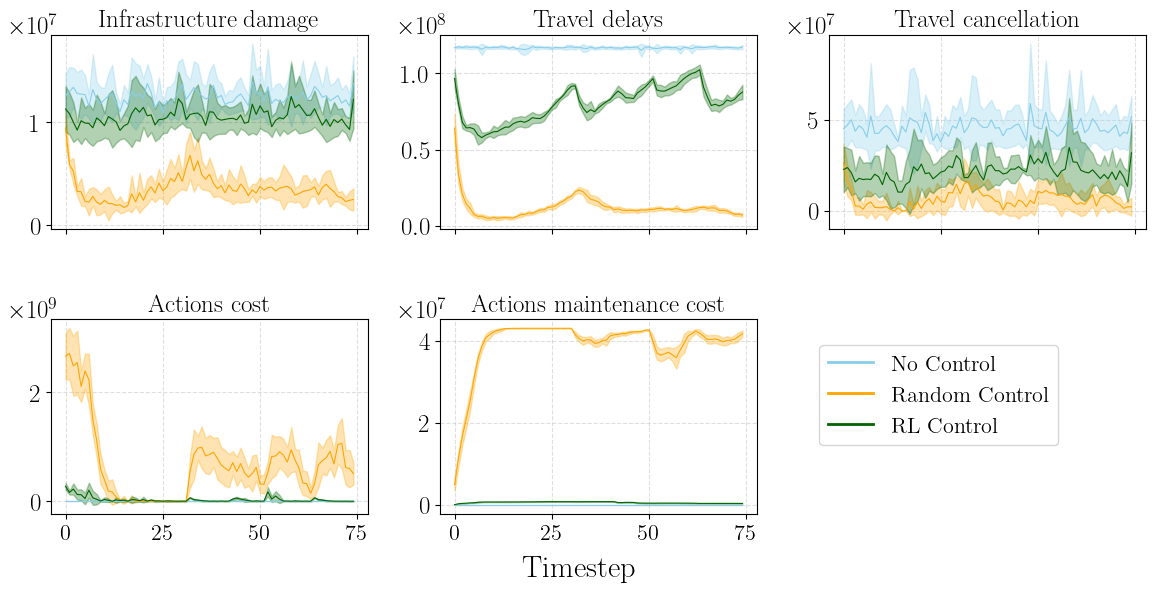} \\
    \vspace{-10pt}
    \caption{Average (and standard deviations shaded) for all five reward components over the period 2024--2100 under the RCP4.5 climate scenario. While Random Control reduces impacts through high and persistent investment costs, the RL policy achieves sustained impact reduction with substantially lower and more stable adaptation expenditures. All values correspond to Danish Kroner (DKK). Please note the different y-axis scale across subfigures.}
    \label{fig:benchmark_algorithms_episode}
\end{figure*}

\begin{table*}[!ht]
    \centering
    \caption{Total reward and costs per reward component using different climate scenarios (RCP2.6, RCP4.5, and RCP8.5) over different 10 runs. Average results are reported ± standard deviations.}
    \label{tab:results_rcps}
    \vspace{3pt}
    \begin{tabular}{lllll}
        \toprule
        \textbf{} & \textbf{RCP2.6} & \textbf{RCP4.5} & \textbf{RCP8.5} \\
        \midrule\midrule
        Infrastructure damage & 8.00 ± 0.18$\times10^8$ & 7.90 ± 0.13$\times10^8$ & 8.05 ± 0.19$\times10^8$ \\
        Travel delays & 6.55 ± 0.10$\times10^9$ & 6.01 ± 0.06$\times10^9$ & 5.54 ± 0.17$\times10^9$ \\
        Travel cancellations & 1.63 ± 0.10$\times10^9$ & 1.61 ± 0.09$\times10^9$ & 1.49 ± 0.10$\times10^9$ \\
        Action costs & 1.73 ± 0.09$\times10^9$ & 2.32 ± 0.11$\times10^9$ & 3.27 ± 0.34$\times10^9$ \\
        Action maintenance costs & 3.46 ± 0.32$\times10^7$ & 4.23 ± 0.21$\times10^7$ & 1.88 ± 0.57$\times10^8$ \\
        \midrule
        Total reward\footnotemark & -1.07 ± 0.02$\times10^{11}$ & -1.08 ± 0.01$\times10^{11}$ & -1.13 ± 0.02$\times10^{11}$ \\
        \bottomrule
    \end{tabular}
\end{table*}

\subsection{Reinforcement learning for climate adaptation planning}
We benchmark the learned RL policy against two computationally feasible baselines under RCP4.5: \textit{No Control (NC)}, which never deploys measures, and \textit{Random Control (RND)}, which selects actions uniformly at random across zones and time. Figure~\ref{fig:benchmark_algorithms} reports the cumulative objective and its decomposition into the five cost components (infrastructure damage, delays, cancellations, action costs, and maintenance). The RL policy achieves the best overall performance, reducing the cumulative total cost by 22\% relative to NC and by 408\% relative to RND. While NC and RND are intentionally simple references, the comparison highlights the value of coordinated long-horizon decisions: RND can reduce some impact components, but does so via excessive and poorly timed spending, which dominates the objective.

Figure~\ref{fig:benchmark_algorithms_episode} shows the corresponding temporal profiles. NC exhibits persistently high impact costs over the horizon. RND reduces impacts early, but sustains large action and maintenance costs throughout the episode, indicating redundant or mistimed deployments. In contrast, the RL policy learns a more targeted investment pattern—incurring substantially lower and more stable expenditures while achieving sustained reductions in infrastructure damage, delays, and cancellations. Overall, the learned policy operationalizes the intended trade-off between upfront investment and avoided long-run losses.

\footnotetext{Notice that total reward is given by Eq.~\ref{eq:reward} (i.e., $-\sum_t r_t$), thus costs have the opposite sign of rewards.}
\begin{table}
    \centering
    \caption{Total reward using an RL strategy that was designed using various climate beliefs and reality scenario pairs. Average results are reported ± standard deviations.}
    \label{tab:train_one_scenario_test_another}
    \vspace{3pt}
    \begin{tabular}{llr}
    \toprule
    Belief Scenario & Reality Scenario & Reward \\
    \midrule
    \midrule
    RCP2.6 & RCP2.6 & -107.42 ± 1.55 \\
           & RCP4.5 & -107.53 ± 1.54 \\
           & RCP8.5 & -109.94 ± 1.19 \\
    RCP4.5 & RCP2.6 & -107.42 ± 1.55 \\
           & RCP4.5 & -107.67 ± 1.29 \\
           & RCP8.5 & -109.45 ± 1.15 \\
    RCP8.5 & RCP2.6 & -110.18 ± 1.27 \\
           & RCP4.5 & -110.66 ± 0.90 \\
           & RCP8.5 & -113.03 ± 1.78 \\
    \bottomrule
\end{tabular}

\end{table}

\subsection{Performance across climate scenarios and robustness}
Table~\ref{tab:results_rcps} summarizes performance when training and evaluating the RL policy within each climate scenario (RCP2.6/4.5/8.5). Across scenarios, the policy increases adaptation spending as projected conditions become more severe (higher action and maintenance costs under RCP8.5), while keeping the monetized transport impact components comparatively contained. In fact, it turns out that, when trained on a rather pessimistic scenario (RCP8.5), the increase in action spending is strongly compensated by the savings in impacts.

 To further assess the value of \emph{pessimistic} vs. \emph{optimistic} priors, we run counterfactual \emph{belief–reality} tests: train under scenario A (belief) and evaluate under scenario B (realized).
Table~\ref{tab:train_one_scenario_test_another} reports cumulative objective values \textbf{in units of $10^{9}$ DKK} (mean $\pm$ std; lower is better).
As expected, outcomes are best when the realized climate is mild (RCP2.6) and deteriorate for harsher realizations (RCP8.5), indicating under-adaptation when training assumes mild futures.
Averaged across realizations, the RCP4.5-trained policy performs best, suggesting an intermediate belief balances performance and robustness.
Policies trained under more severe beliefs sacrifice some surplus in mild realizations but lose less in harsh ones, which is consistent with precautionary investment carrying an opportunity cost.

Taken together, these results illustrate a performance-robustness trade-off and show how the framework can explicitly quantify the consequences of incorrect climate assumptions in long-horizon adaptation planning.
% ==================================================== %
% Discussion
% ==================================================== %
\section{Discussion}
\label{sec:discussion}

\subsection{Stakeholders and practical impact}
This work is part of long-running collaboration with Copenhagen, Esbjerg and Odense Municipalities, as well as the Paris region transport operator (RATP), with the goal of supporting long-horizon adaptation planning where decisions are sequential, spatially coupled, and evaluated under deep climate uncertainty. The framework is intended as a \emph{decision-support} tool: it enables systematic stress-testing of multiple and alternative intervention portfolios, quantifies trade-offs between investment and avoided transport disruption, and produces interpretable spatial-temporal adaptation pathways that can be compared against planning priorities and constraints. Beyond the Copenhagen case study here demonstrated, the same IAM+RL structure is being instantiated with alternative hazard (i.e., other weather event types), impact (e.g., other urban systems beyond transport), or valuation modules (e.g., private and social costs, health-related impacts), in the above-mentioned cities.

\subsection{Methodological implications}
\begin{figure*}[!htb]
    \centering
    \includegraphics[width=.98\textwidth]{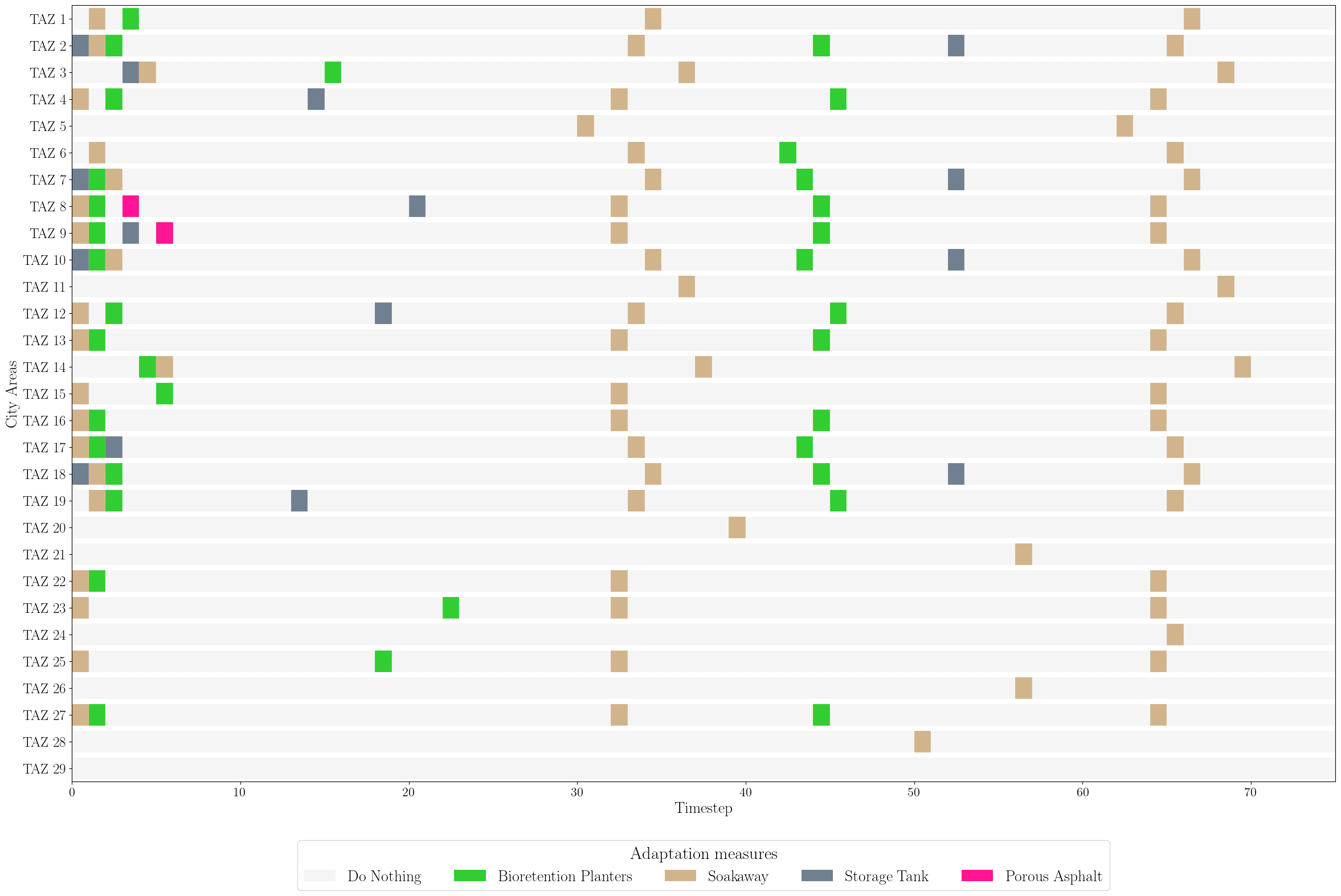} \\
    \caption{Adaptation measures taken over time per zone in Copenhagen's city center for one episode (run) for a RCP4.5 scenario.}
    \label{fig:adaptation_pathway}
\end{figure*}

Methodologically, the results show that reinforcement learning can learn non-trivial, long-horizon policies in a coupled hazard--transport setting with high-dimensional spatial state and delayed consequences. The learned strategies differ from simplistic baselines by jointly timing and placing interventions, balancing upfront investment and maintenance against avoided damage, delay, and cancellation costs. To illustrate this point, Figure \ref{fig:adaptation_pathway} shows an example of a possible adaptation pathway for Copenhagen, across time. Importantly, however, the framework should be interpreted as an exploratory engine for generating and evaluating \emph{collections} of adaptive pathways, rather than as a prescriptive optimizer that outputs a single “best” plan for a single scenario. In practice, we can provide a distribution of such pathways that can be further inspected in search for global robustness plans (e.g. "implementing action $A$ before time $t$ in zone $Z$ is be optimal in $X\%$ of the analysed scenarios"). 

\subsection{Limitations and ethics}
Several limitations are inherent to simulation-based decision support. First, outcomes depend on modeling assumptions in the hazard, transport, and valuation modules; results should therefore be read as conditional on the modeled dynamics rather than as forecasts. Second, our climate uncertainty treatment uses three discrete scenarios and does not represent probabilistic trajectories or belief updating over time. Third, training remains computationally intensive, constraining the size of the action space and the number of scenarios that can be explored.

From an ethics and governance perspective, the framework optimizes a monetized objective and may omit distributional impacts (e.g., equity, accessibility for vulnerable groups) unless explicitly modeled. Any practical use should therefore incorporate additional objectives and constraints, and be embedded in a stakeholder process to ensure transparency, accountability, and alignment with public values.

\section{Conclusions}
\label{sec:conclusions}
We introduced an integrated IAM+RL framework for long-term adaptation planning that couples climate-driven flood hazard modeling, transport simulation, and monetized impact accounting. In a case study of Copenhagen’s inner city (2024--2100), the learned policies achieve lower cumulative costs than no-adaptation and random baselines by producing coordinated spatial-temporal pathways and balancing investment/maintenance costs against avoided disruptions. Cross-scenario experiments further illustrate how performance changes under climate uncertainty and enable explicit robustness analysis.

Future work will extend the framework to probabilistic climate ensembles and belief updating, multi-objective formulations capturing other social impacts (wellbeing and equity), and computational acceleration (e.g., surrogate components) to scale to larger cities and richer intervention sets.

\bibliographystyle{named}
\bibliography{references}

\end{document}